\makeatletter
\newcommand{\dontusepackage}[2][]{%
  \@namedef{ver@#2.sty}{9999/12/31}%
  \@namedef{opt@#2.sty}{#1}}
\makeatother
\dontusepackage{subfigure}

\documentclass[]{article}
\usepackage{bbm}
\usepackage{lmodern}
\usepackage{amssymb,amsmath}
\usepackage{ifxetex,ifluatex}
\usepackage[usenames,dvipsnames]{color}
\usepackage{fixltx2e} 
\ifnum 0\ifxetex 1\fi\ifluatex 1\fi=0 
  \usepackage[T1]{fontenc}
  \usepackage[utf8]{inputenc}
\else 
  \ifxetex
    \usepackage{mathspec}
    \usepackage{xltxtra,xunicode}
  \else
    \usepackage{fontspec}
  \fi
  \defaultfontfeatures{Mapping=tex-text,Scale=MatchLowercase}
  
\fi
\IfFileExists{upquote.sty}{\usepackage{upquote}}{}
\IfFileExists{microtype.sty}{%
\usepackage{microtype}
\UseMicrotypeSet[protrusion]{basicmath} 
}{}
\usepackage[margin=1.0in,bottom=1.5in]{geometry}
\usepackage[square,numbers,sort&compress]{natbib}
\usepackage{listings}
\usepackage{graphicx}
\makeatletter
\def\maxwidth{\ifdim\Gin@nat@width>\linewidth\linewidth\else\Gin@nat@width\fi}
\def\maxheight{\ifdim\Gin@nat@height>\textheight\textheight\else\Gin@nat@height\fi}
\makeatother
\setkeys{Gin}{width=\maxwidth,height=\maxheight,keepaspectratio}
\usepackage{caption}
\usepackage{float}
\usepackage{subfig}
\ifxetex
  \usepackage[setpagesize=false, 
              unicode=false, 
              xetex]{hyperref}
\else
  \usepackage[unicode=true]{hyperref}
\fi
\hypersetup{breaklinks=true,
            bookmarks=true,
            pdfauthor={},
            pdftitle={Practical Bayesian experimental design with conditional normalizing flows},
            colorlinks=true,
            citecolor=black,
            urlcolor=blue,
            linkcolor=black,
            pdfborder={0 0 0}}
\usepackage[preprint]{neurips_2019}
\usepackage{url}
\usepackage{booktabs}
\usepackage{amsfonts}
\usepackage{nicefrac}
\usepackage{microtype}

\makeatletter
\let\@oldfnsymbol\@fnsymbol
\renewcommand{\@fnsymbol}[1]{\@oldfnsymbol{0}}
\makeatother

\title{Probabilistic Bayesian optimal experimental design using conditional normalizing flows}
\author{
    Rafael Orozco\textsuperscript{1*}\thanks{* Correspondence: \href{mailto:rorozco@gatech.edu}{rorozco@gatech.edu}},
    Felix J. Herrmann\textsuperscript{1},
    Peng Chen \textsuperscript{1}\\
    \textsuperscript{1}Georgia Institute of Technology
}
\date{}

\begin{document}
\maketitle
\begin{abstract}
Bayesian optimal experimental design (OED) seeks to conduct the most informative experiment under budget constraints to update the prior knowledge of a system to its posterior from the experimental data in a Bayesian framework. Such problems are  computationally challenging because of (1) expensive and repeated evaluation of some optimality criterion that typically involves a double integration with respect to both the system parameters and the experimental data, (2) suffering from the curse-of-dimensionality when the system parameters and design variables are high-dimensional, (3) the optimization is combinatorial and highly non-convex if the design variables are binary, often leading to non-robust designs. To make the solution of the Bayesian OED problem efficient, scalable, and robust for practical applications, we propose a novel joint optimization approach. This approach performs simultaneous (1) training of a scalable conditional normalizing flow (CNF) to efficiently maximize the expected information gain (EIG) of a jointly learned experimental design (2) optimization of a probabilistic formulation of the binary experimental design with a Bernoulli distribution. 
We demonstrate the performance of our proposed method for a practical MRI data acquisition problem, one of the most challenging Bayesian OED problems that has high-dimensional (320 $\times$ 320) parameters at high image resolution, high-dimensional (640 $\times$ 386) observations, and binary mask designs to select the most informative observations. 
\end{abstract}

\section{Introduction}\label{introduction}


When solving an inverse problem, the goal of experimental design is to chose how to observe data from the field that is used to infer an unknown parameter. The process that models the data acquisition is denoted as the forward process written 
$$
\mathbf{y} = \mathbf{M}(\mathcal{F}(\mathbf{x})) + \boldsymbol{\varepsilon}
$$
where $\mathcal{F}(\cdot)$ is the forward operator that takes the unknown $\mathbf{x}$ to the observation space,  $\mathbf{M}(\cdot)$ is the observation process that need not be linear and $\boldsymbol{\varepsilon}$ is corruption noise. Since experimentalists have control over the observation process, it is desirable to control this in order to best inform downstream inferences of $\mathbf{x}$ i.e experimental design. As an illustration, imagine a doctor deciding where to place a handheld ultrasound on a patient to best infer the state of a internal organ. 

In a Bayesian framework, the experimental design is based on quantities related to the posterior distribution $p(\mathbf{x}|\mathbf{y})$. Once the posterior has been estimated, different design optimality choices can be made. These options include A-optimal, D-optimal, and the expected information gain ($EIG$) \cite{lindley1956measure}.
Due to its close connection with the posterior likelihood (the quantity used to train normalizing flows) we focus on the EIG: 
$$
EIG(\mathbf{M}) = \mathbb{E}_{p(\mathbf{y}|\mathbf{M})}\,\,\left[ D_{KL} (p(\mathbf{x}|\mathbf{y},\mathbf{M})\, || \, p(\mathbf{x})) \right]
$$
which measures the information gain between the prior information and the information gained by performing an experiment using $\mathbf{M}$ to acquire data $\mathbf{y}$. Importantly, the expectation $\mathbb{E}_{p(\mathbf{y}|\mathbf{M})}$ means that the information gain is averaged over the distribution of possible observations i.e it describes the best experiment on average for the range of observations that is expected to be encountered. However, optimizing EIG is challenging due to complexities like the "double intractability" problem, which arises from the necessity to evaluate two expectations \cite{foster2019variational}. Our approach tackles this with a data-driven optimization of the EIG that combines simulation based inference \cite{cranmer2020frontier}, likelihood-based generative models \cite{dinh2014nice} to tractably find optimal designs and a probablistic interpretation of the design parameters to facilitate its optimization.

\section{Methodology}\label{methodology}
To demonstrate our scalable technique for Bayesian experimental design, we first show the connection between $EIG$ and likelihood based generative models. Secondly, we present conditional normalizing flows as the key tool due to their exact likelihood evaluation and their invertible architectures that enable memory efficient training. Then we show how a probabilistic interpretation of binary design masks alleviates optimization challenges. Finally, we setup a demonstration of our technique as applied to high dimensional inverse problem related to MRI medical imaging. 

\subsection{Normalizing flows learn the expected information gain}
In the context of summary statistics, \cite{hoffmann2022minimizing} noted that a summary statistic $\mathbf{\bar y}$ can be interpreted as the transformation on observations $\mathbf{y}$ that both maximizes the posterior likelihood $p(\mathbf{x}| \mathbf{\bar y})$, and also maximizes the expected information gain $EIG(\mathbf{\bar y})$. By interpreting the result from an experiment $\mathbf{M}$ as a summary statistic, we use a similar derivation to show that maximizing the $EIG$ of a design $\mathbf{M}$ is equivalent to maximizing the posterior likelihood that the design induces in expectation over a joint probability $p(\mathbf{x},\mathbf{y}|\mathbf{\mathbf{M}})$:
\begin{align}
\underset{\mathbf{M}}{\operatorname{max}}
 \,\,\,EIG(\mathbf{M}) =& \mathbb{E}_{p(\mathbf{y}|\mathbf{M})}\,\,\left[ D_{KL} (p(\mathbf{x}|\mathbf{y})\, || \, p(\mathbf{x})) \right] \\
 =& \mathbb{E}_{p(\mathbf{y}|\mathbf{M})} \,\, \left[  \,\, \mathbb{E}_{p(\mathbf{x}|\mathbf{y})} \left[  \log p(\mathbf{x}|\mathbf{y}) - \log  p(\mathbf{x}) \right]\right] \\
  =& \mathbb{E}_{p(\mathbf{y}|\mathbf{M})} \,\, \left[  \,\, \mathbb{E}_{p(\mathbf{x}|\mathbf{y})} \left[  \log p(\mathbf{x}|\mathbf{y}) \right]\right] \\
  =& \mathbb{E}_{p{}(\mathbf{x},\mathbf{y}|\mathbf{M})} \,\, \left[   \log p(\mathbf{x}|\mathbf{y})\right].
 \end{align} 
 We have simplified the expressions by using the fact that the maximization is constant with respect to the prior $p(\mathbf{x})$ in line $3$ and then the law of total expectation in the last line. Crucially, the final expression is equivalent to the quantity (posterior log likelihood) that is used to guide optimization of likelihood-based conditional generative models:
\begin{equation} \label{eq:opt-cnf}
   \underset{\theta}{\operatorname{max}} \, \mathbb{E}_{p{}(\mathbf{x},\mathbf{y})} \left[ \log p_{\theta}(\mathbf{x}|\mathbf{y}) \right],
\end{equation} 
where $\theta$ are the network weights that define the conditional generative model. The equivalence between the $EIG$ and the objective in Equation \ref{eq:opt-cnf} implies that we can setup a joint optimization:
\begin{equation} \label{eq:opt-cnf}
   \underset{\theta, \mathbf{M}}{\operatorname{max}} \, \mathbb{E}_{p{}(\mathbf{x},\mathbf{y}|\mathbf{M})} \,\, \left[   \log p_{\theta}(\mathbf{x}|\mathbf{y})\right].
\end{equation} 
and that the gradient signal from the objective can be back-propagated to the design $\mathbf{M}$ to calculate an update $\nabla_{\mathbf{M}}$ and it will point in the direction of increased expected information gain. Using conditional generative models for EIG optimization has been previously suggested by separate arguments in \cite{foster2020unified}.  A particular class of models that is trained with this objective are conditional normalizing flows \cite{radev2020bayesflow,ardizzone2019guided} a type of generative model that is known to be universal approximators of distributions \cite{draxler2024universality}. Conditional normalizing flows are an attractive choice because of their exact likelihood evaluations that enables the aforementioned joint optimization in the first place and also because they are invertible by design thus lead to efficient memory use during training on large inputs. In this work, we exploit this equivalence and conditional normalizing flows to perform Bayesian optimal experimental design on a large-scale imaging problems. Next we meet the challenge of finding binary designs by reinterpreting the design as a probabilistic sampling pattern.

\subsection{Probabilistic mask design}
We express the design parameters as binary mask where a $1$ means that we have placed a sensor at this location and $0$ otherwise. Instead of directly optimizing for binary mask $\mathbf{M}$, we optimize for parameters of a Bernoulli distribution for each possible binary value. This is achieved by following the methods from \cite{bahadir2019learning} where we parameterize the distribution as real values $\mathbf{w}$ then create a binary the mask by applying the indicator function  $\mathbbm{1}_{\mathbf{w} < \mathbf{u}}$ where $\mathbf{u}$ is sampled from the uniform distribution $\mathbf{u} \sim U(0,1)$. We insure that the sampling budget is respected by normalizing $\mathbf{w}$ such that the average value is kept equal to the budget $s$. Thus the binary mask is defined as
\begin{align} \label{eq:binarize}
 \mathbf{M}(\mathbf{w}) := \mathbbm{1}_{\text{s}\frac{\mathbf{w}}{\| \mathbf{w} \|}  < \mathbf{u} } \\ 
 \text{where} \, \, \, \mathbf{u} \sim U(0,1).
\end{align}
When optimizing for $\mathbf{w}$ with gradient descent, we calculate the gradients of the indicator function with the pass-through gradient approximation from \cite{bengio2013estimating} by treating the indicator function as the identity during back propagation. We chose to express the binary mask as in a probabilistic manner for two reasons. First, probabilistic sampling of the mask during training aids in jumping out of local minima that would be challenging to avoid if the mask was deterministic. Secondly, because the optimized mask represents a relative likelihood of sampling then the sampling budget $s$ can be changed post-hoc by the user. In other words, changing the sampling budget does not require retraining the network instead requires simple a re-scaling of the learned probabilistic mask $\mathbf{w}$.

\subsection{Experimental design for high-dimensional medical imaging}
Magnetic Resonance Imaging (MRI) is an important modality that is routinely used in the diagnosis of diseases related to cancer, neurology and the musculoskeletal system. The measured field by an MRI machine is spatial frequencies of the patient tissue. Each measurement takes time thus taking less measurements translates to savings in money due to expensive operation and also results in increased patient comfort.

We use the FAST MRI knees dataset \cite{zbontar2018fastmri} to create a training pairs comprised of ($320\times320$) ground truth multi-coil images and ($640\times386$) single-coil k-space observations as complex numbers. We use $1800$ training samples (a relatively low number as compared to related work \cite{bahadir2019learning,yin2021end} that used between 17k-34k training pairs.) 


Given the Fourier transformation operator $\mathbf{A}$, we solve the following maximization with stochastic gradient descent to jointly train a conditional normalizing flow $f_{\theta}$ and the probabilistic design parameterized by $\mathbf{w}$:

\begin{equation} \label{eq:opt-joint}
    \hat \theta,  \mathbf{\hat w} = \underset{\theta,\, \mathbf{w}}{\operatorname{arg max}} \,  \frac{1}{N} \sum_{i=1}^{N}  \left( -\frac{1}{2}\lVert  f_{\theta}(\mathbf{x}^{(i)}; \mathbf{A}^{\top} \mathbf{M}(\mathbf{w}) \odot \mathbf{ y}^{(i)}) \rVert_{2}^2 + \log \left|  \det{  \mathbf{J}_{f_{\theta}}} \right| \right),
\end{equation} 
where $\mathbf{A}^{\top}$ is the adjoint Fourier transform, and $\mathbf{J}_{f_{\theta}}$ is the Jacobian of the normalizing flow as is 
needed for maximum-likelihood training as per the change of variables formula. Here we implement our conditional normalizing flow with InvertibleNetworks.jl \cite{orozco2023invertiblenetworks} that takes advantage of the invertiblity of normalizing flow layers to achieve low-memory requirements during training. After training, the network is an amortized sampler for the posterior distribution for all observations in the training distribution. Given the optimized network parameters and the optimal design we sample from the posterior distribution 
by first acquiring data $\mathbf{y}^{obs}$ from the field as prescribed by the 
optimal design $\mathbf{\hat M}( \mathbf{\hat w})$ and then generating posterior samples by passing Gaussian noise through the inverse network as such: 
\begin{align}
 \mathbf{x}=f^{-1}_{\hat{\boldsymbol{\theta}}}(\mathbf{z};\mathbf{A}^{\top}\mathbf{ y}^{obs}) \,\, \label{eq-post-sampling}  \\ 
 \text{where} \,\,\mathbf{z}\sim \mathcal{N}(0,\,I).
\end{align} 
Note that we decided to process the sub-sampled data with the adjoint Fourier operator $\mathbf{A}^{\top}$ that incurs an additional computational cost during training for our method. For our training hardware, the baseline took $16$ hours to train and our method took $20$ hours on a single GPU. Although the adjoint operator is used during posterior sampling in Equation \ref{eq-post-sampling}, we only need to calculate it once then an arbitrary number of posterior samples can be generated by resampling the Gaussian noise $\mathbf{z}$.

As far as we understand, this is the first work that explores a scalable solution for Bayesian experimental design in MRI so we build our own baseline that used the exact same density estimator $f_{\theta}$. This allows us to understand the uplift achieved by doing experimental design while controlling for the neural network architectures used and also the fact that our solution provides probabilistic solution. In other words, the baseline solves Equation \ref{eq:opt-joint} but only optimizes the network parameters $\theta$ and uses a fixed mask $\mathbf{M}_{b}$. We hand craft the fixed baseline mask by following the methods in \cite{bahadir2019learning,zhang2020extending} and craft a mask that captures low frequencies and then randomly samples high frequencies as shown in Figure \ref{fig-binary-mask-base}.

\section{Results}\label{results}
After training and testing on the FASTMRI dataset, we compare the  posterior inference when conditioning on data produced by a hand-crafted experiment as compared to our method that is conditioned on data that is measured with our optimal design based on maximized expected information gain. We show that posterior samples from our method are more realistic and lead to increased performance in downstream image reconstruction tasks.

\subsection{Optimized experimental designs}
\begin{figure}
\centering
\subfloat[ Learned probabilistic design  $ \mathbf{\hat w}$]{\includegraphics[width=0.240\hsize]{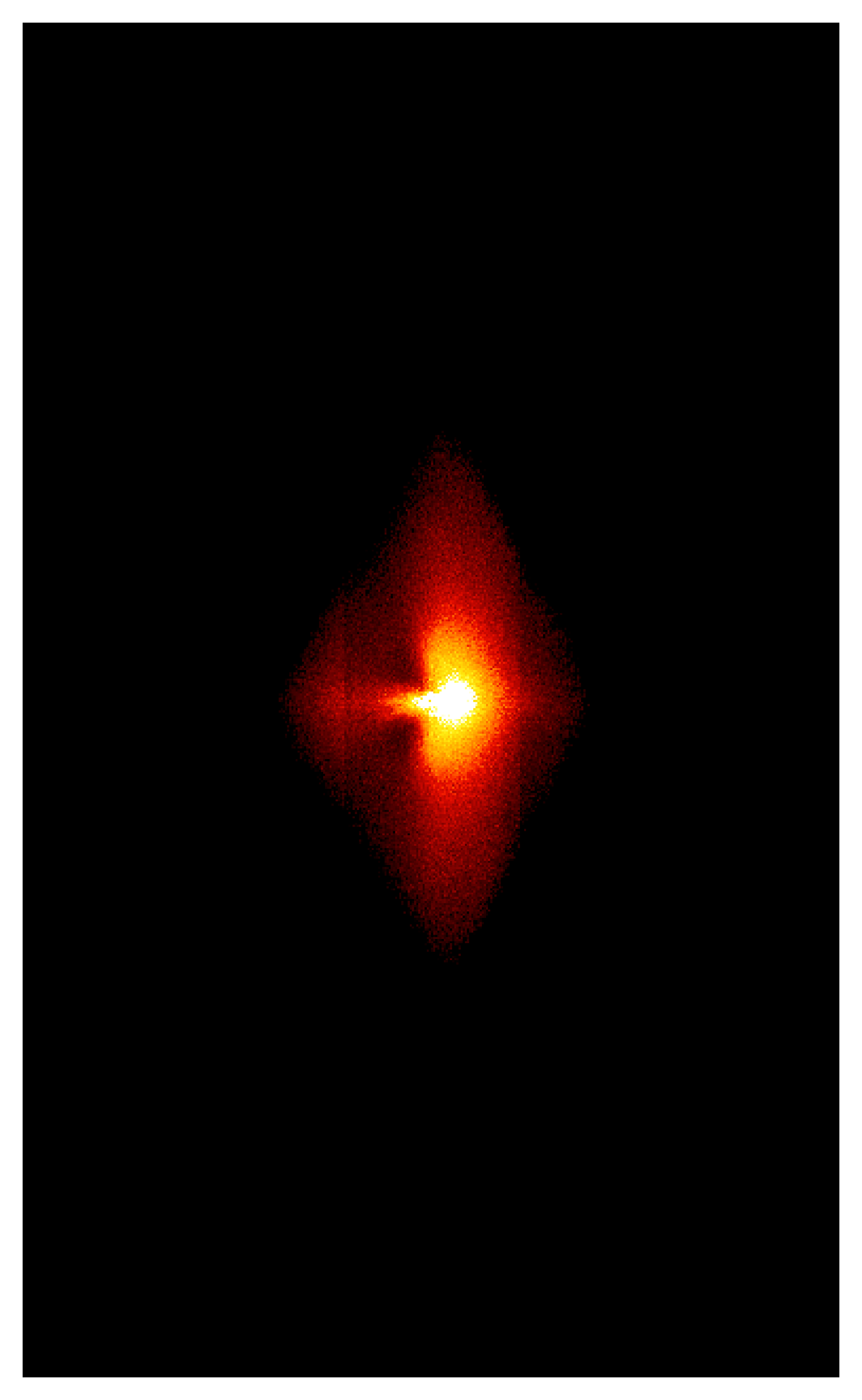}\label{fig-prob-mask}}
\subfloat[ Learned binary design  $\hat {\mathbf{M}}$.]{\includegraphics[width=0.240\hsize]{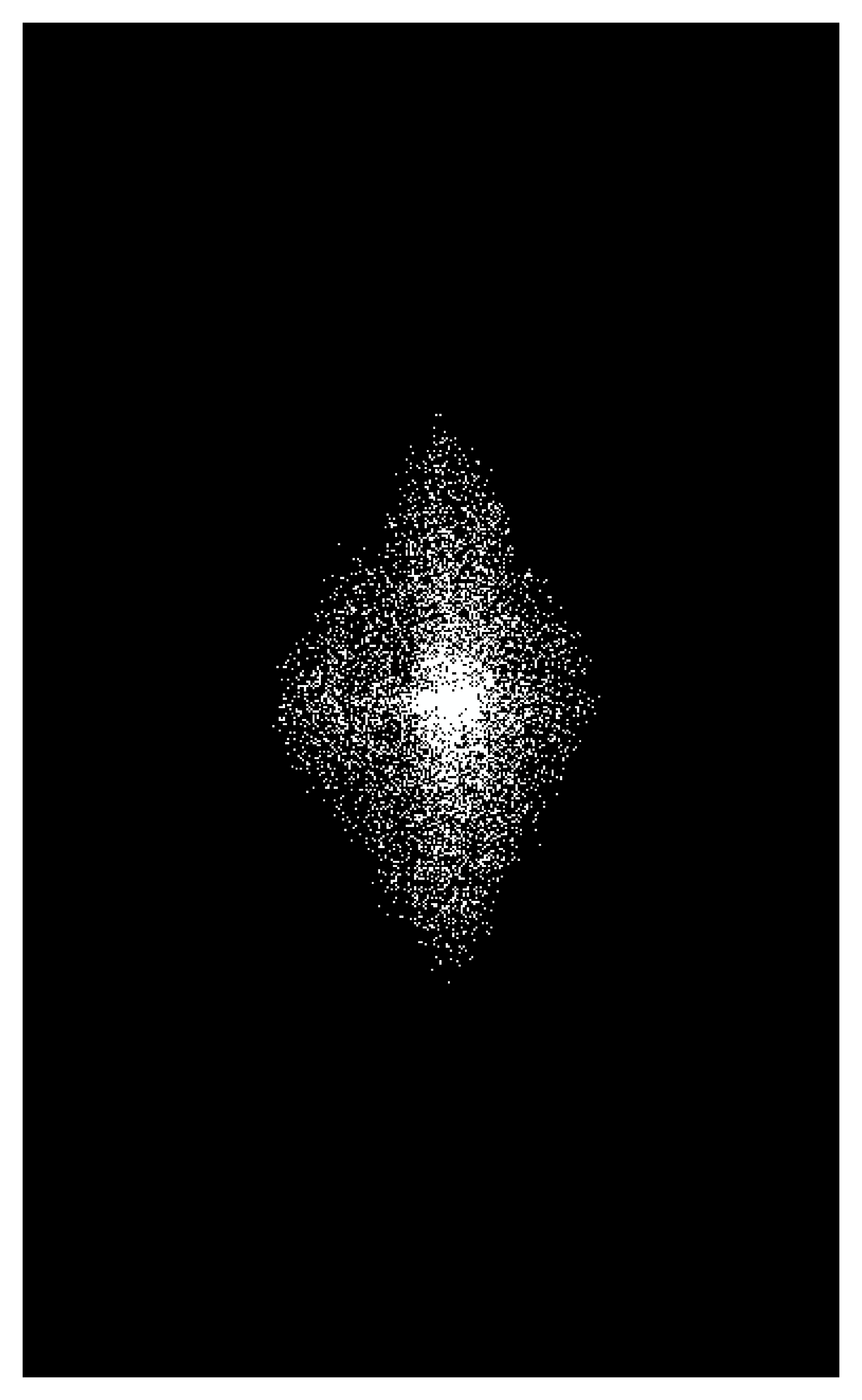}\label{fig-binary-mask}}
\subfloat[ Hand-crafted baseline design $\mathbf{M}_{b}$]{\includegraphics[width=0.240\hsize]{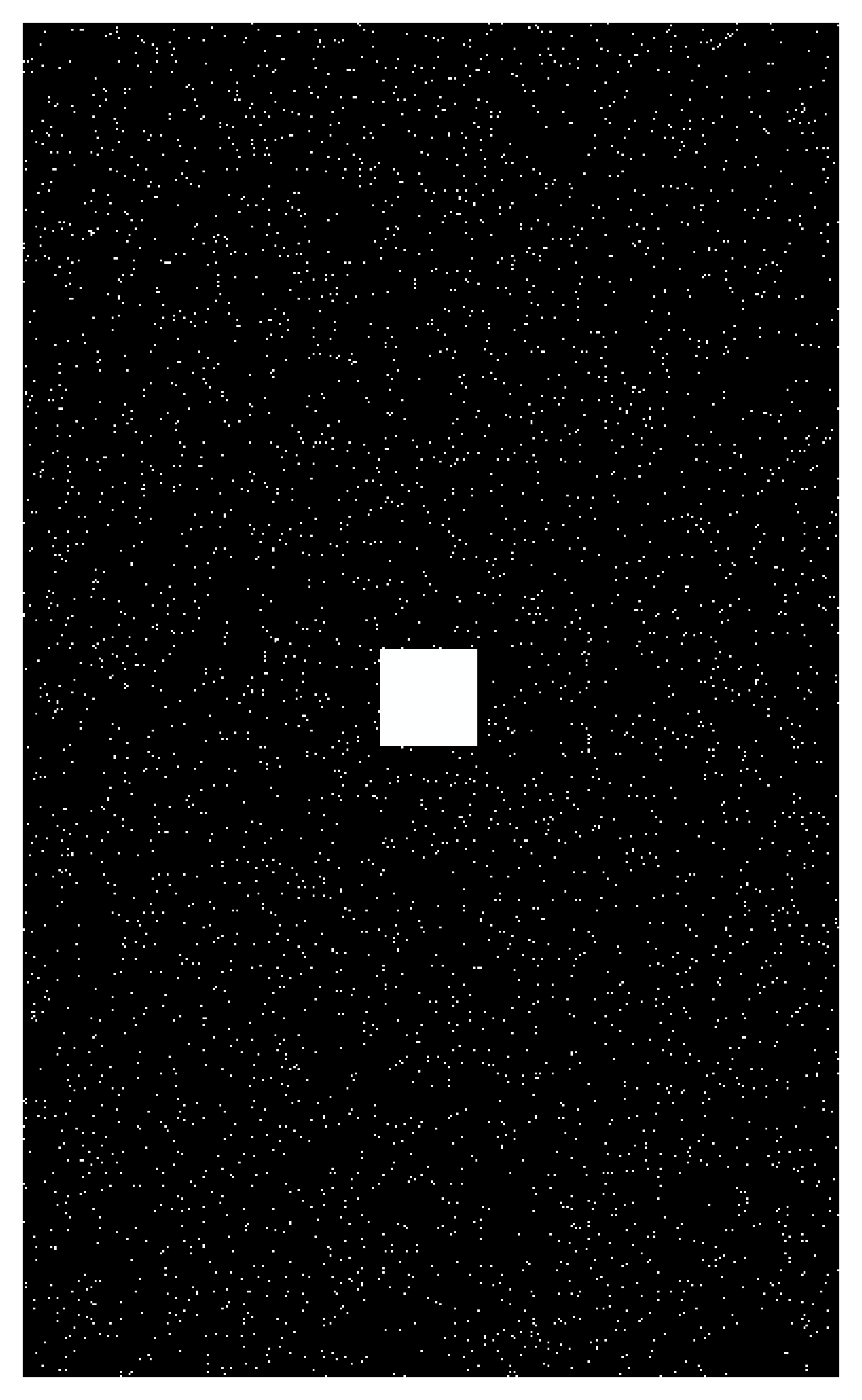}\label{fig-binary-mask-base}} 
\subfloat[ Fully sampled data]{\includegraphics[width=0.240\hsize]{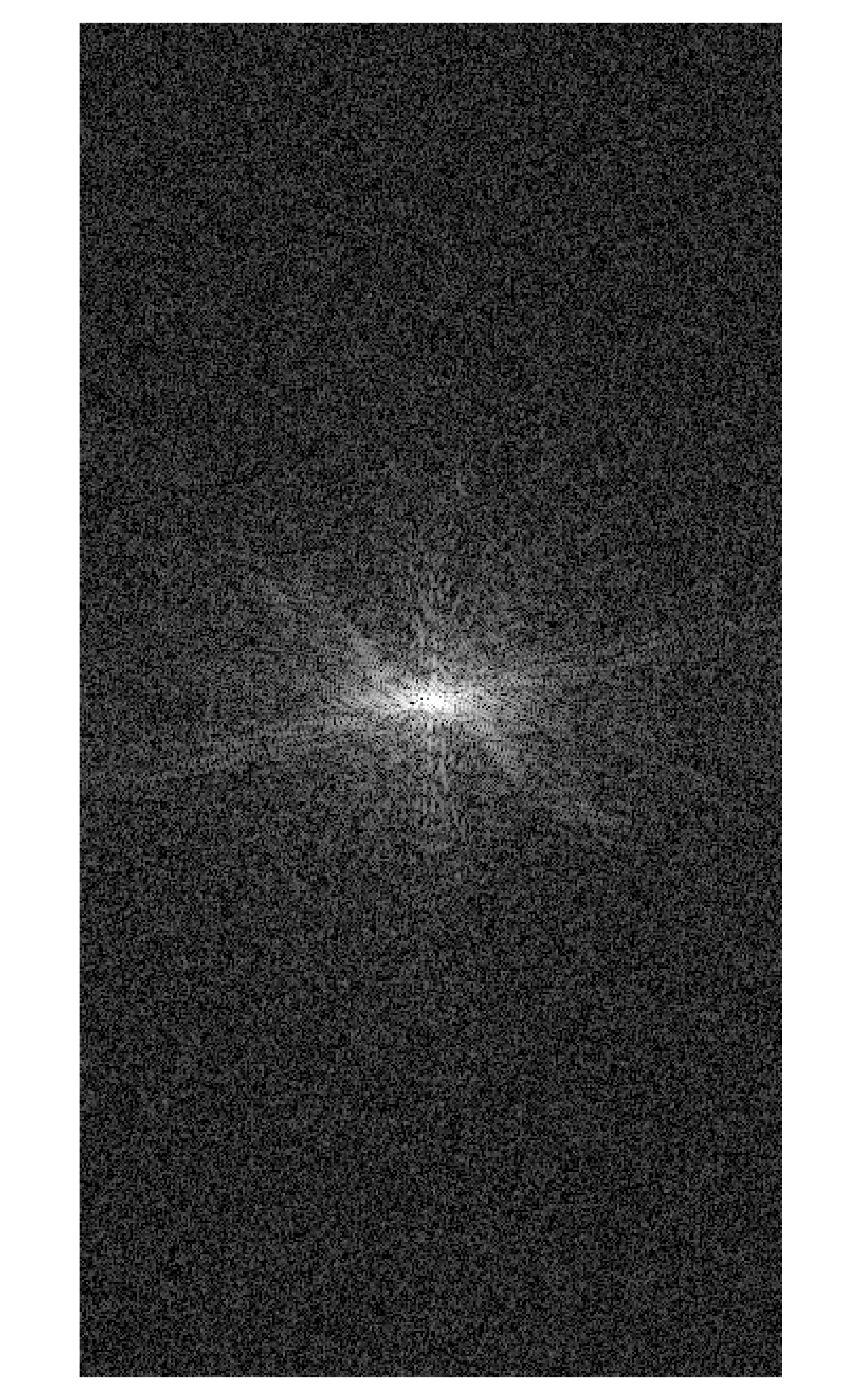}}
\caption{Optimized design compared to hand-crafted design.}\label{fig-masks}
\end{figure}

After training, our method produces two outputs: an amortized posterior sampler $p_{\hat \theta}$ 
and the optimized mask $ \mathbf{\hat M}=\mathbf{M}(\mathbf{\hat w})$. We train and test using a design budget of $2.5\%$ of all k-space frequencies $s=0.025$. Our final optimized probabilistic mask is shown in Figure \ref{fig-prob-mask}, expressed as a sampling density. Using Equation \ref{eq:binarize}, we transform the probabilistic mask to final binary mask that can be used in the field to decide which k-space locations should be sampled to collect data. 

We make the following observations on our optimized design shown in Figure \ref{fig-prob-mask}: \textit{(i)} the optimized design has learned from the training set that it needs to emphasize sampling of low frequencies in the center of the data space but has done so in a smoother more efficient manner than the baseline. \textit{(ii)} the learned design has chosen an anisotropic emphasis on the vertical events in kspace evidenced by the ellipsoid shape of the mask \textit{(iii)} as seen by the asymmetric emphasis of the learned design on the right side of frequency space, the learned design has taken advantage of the Hermitian symmetry that is inherent to MRI machines \cite{wu2017image} without any input of this domain knowledge.

\begin{figure} 
\centering
\captionsetup[subfigure]{labelformat=empty}
\subfloat[(a) Posterior sample $\mathbf{x}\sim p_{\hat \theta }$]{\includegraphics[width=0.30\hsize]{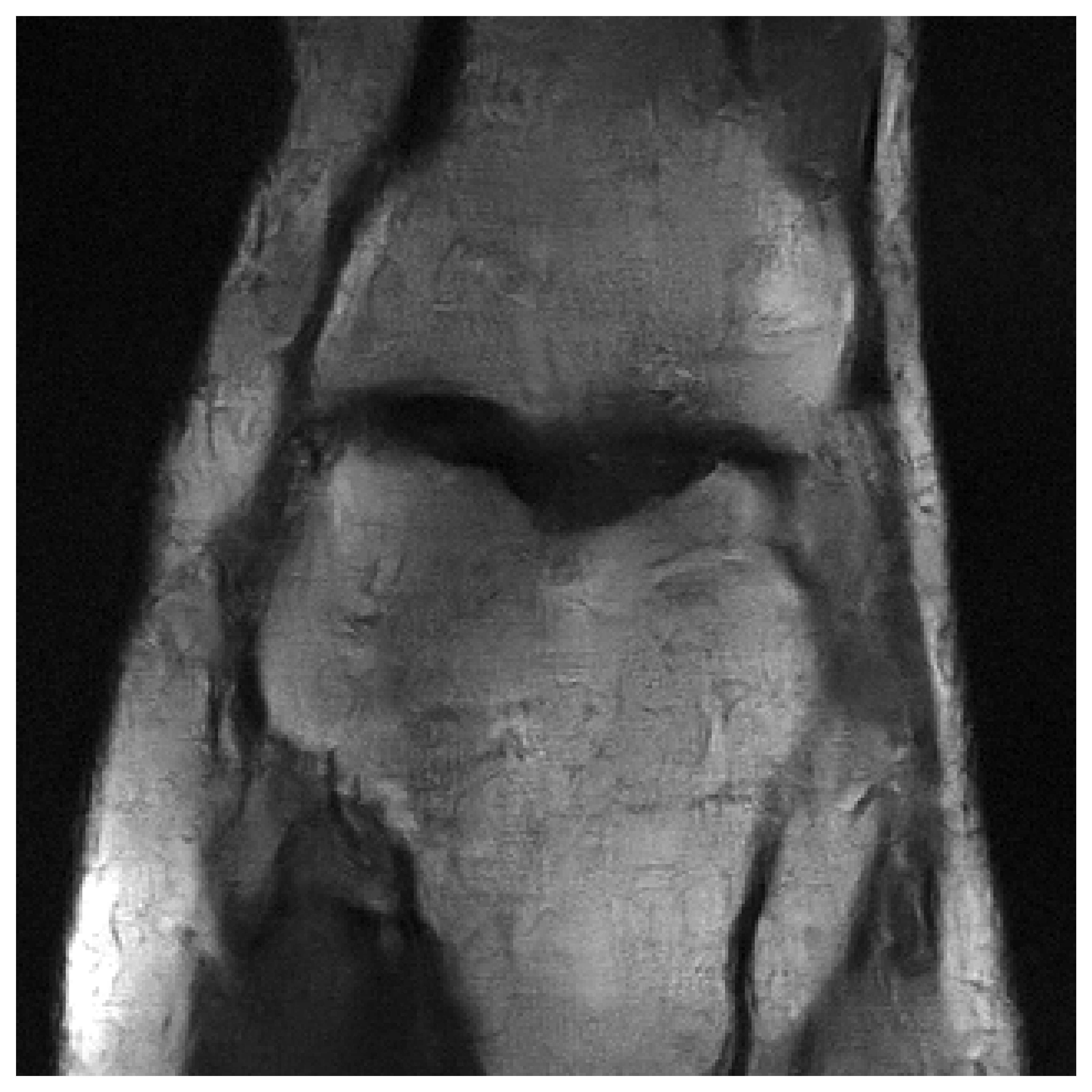}}
\subfloat[(b) Posterior sample $\mathbf{x}\sim p_{\hat \theta }$]{\includegraphics[width=0.30\hsize]{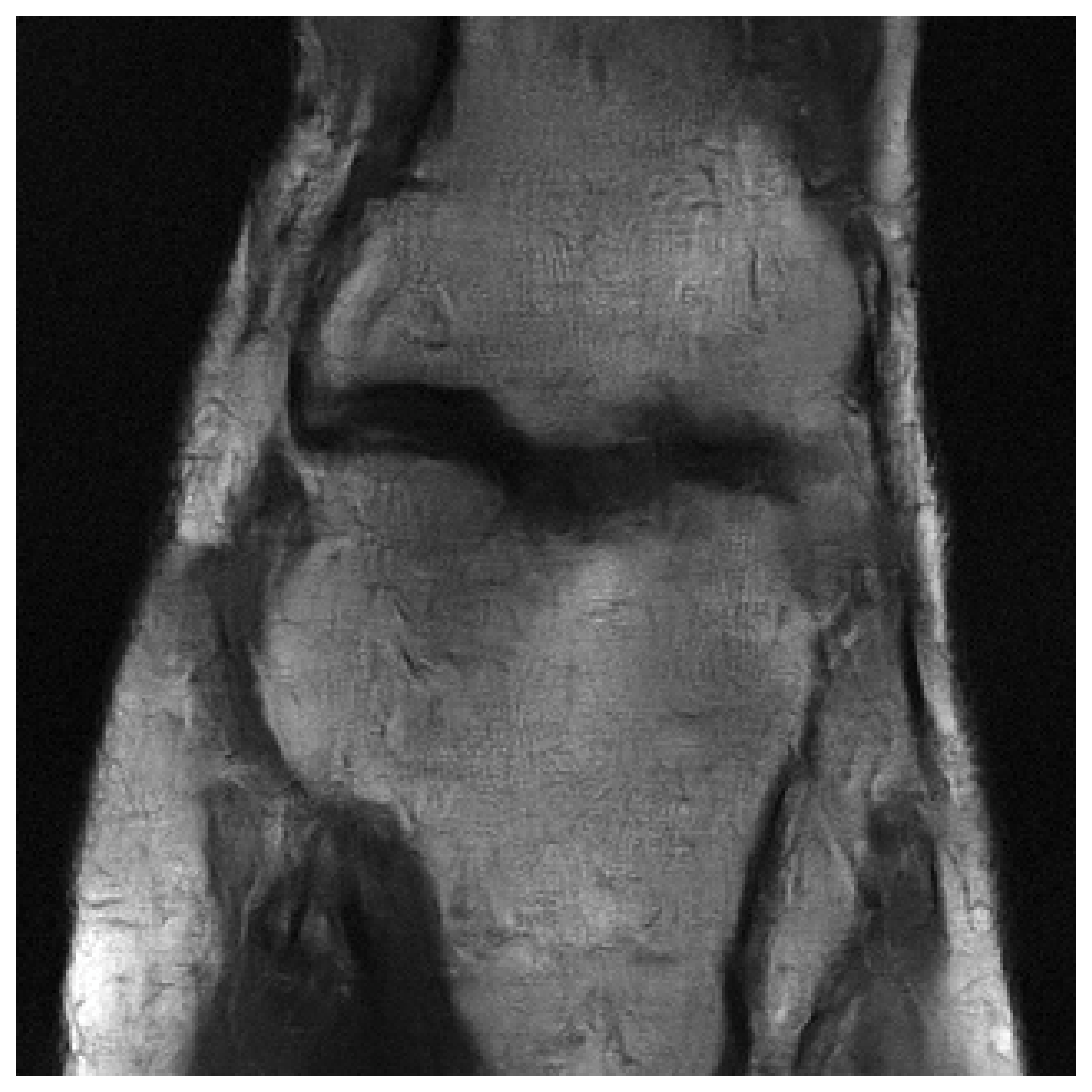}}
\subfloat[(c) Reference image]{\includegraphics[width=0.30\hsize]{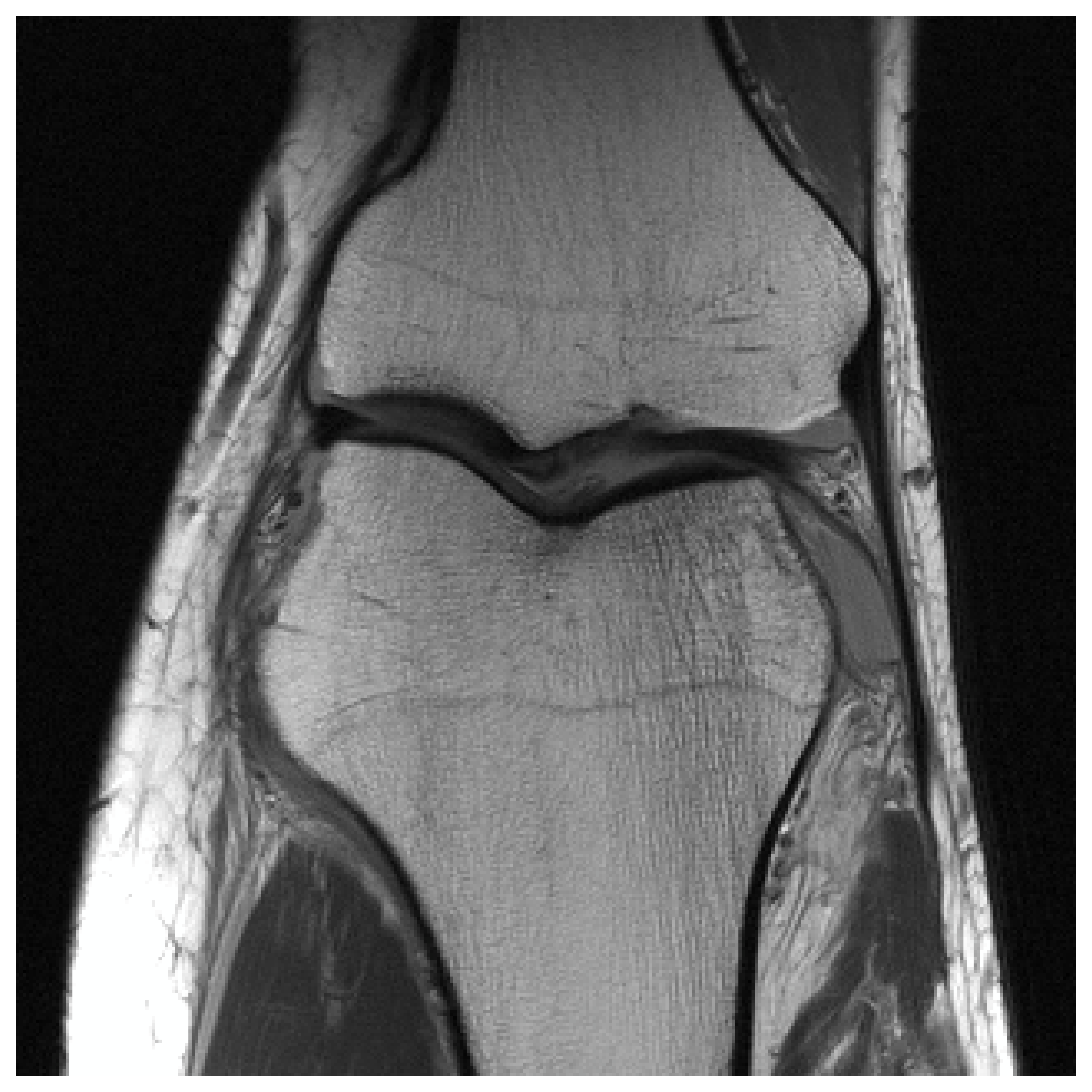}}
\caption{Posterior samples from the baseline method with hand-crafted design.}\label{fig-post-samples-baseline}
\end{figure}

\begin{figure}
\centering
\captionsetup[subfigure]{labelformat=empty}
\subfloat[(a) Our posterior sample $\mathbf{x}\sim p_{\hat \theta  , \hat M }$]{\includegraphics[width=0.30\hsize]{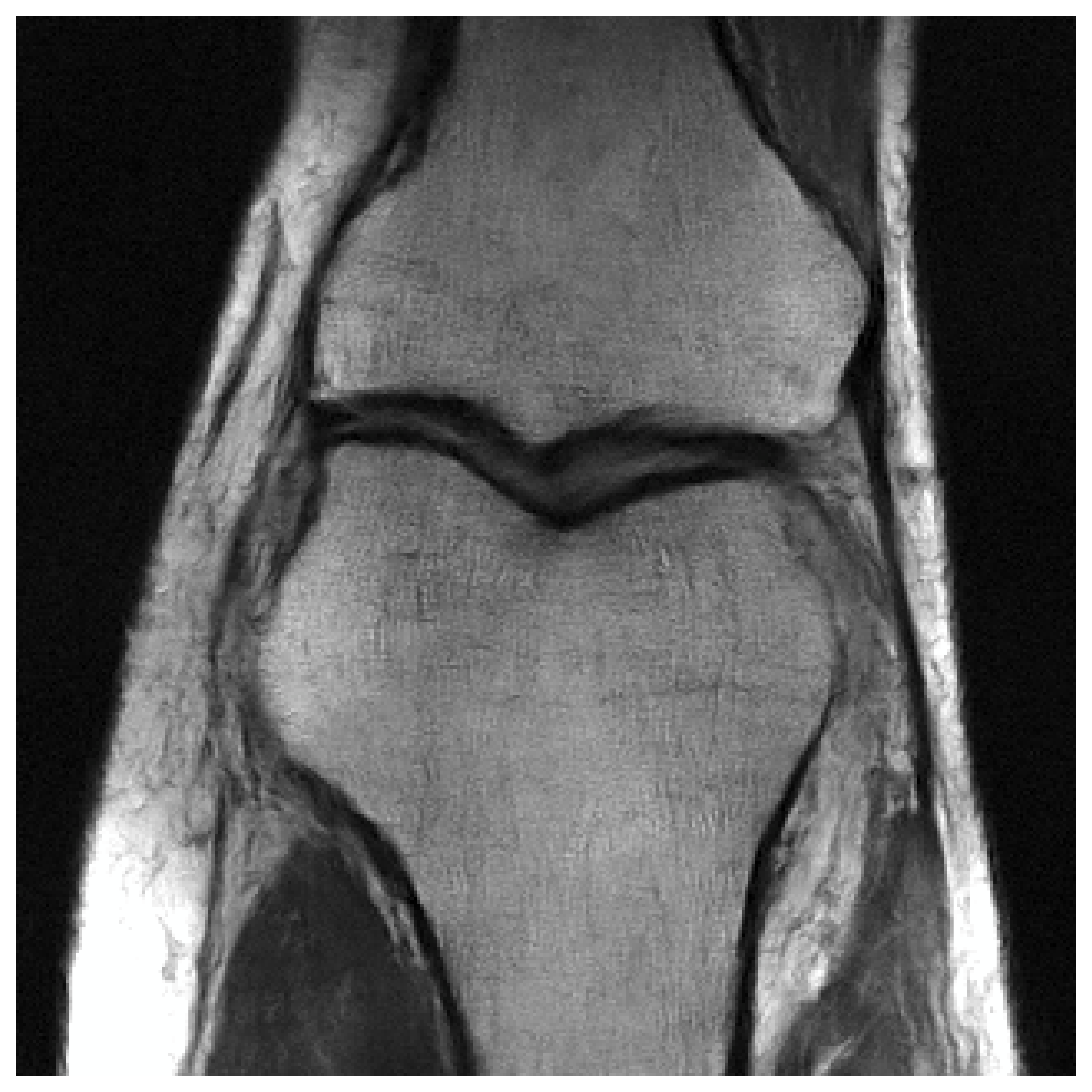}}
\subfloat[(b) Our posterior sample $\mathbf{x}\sim p_{\hat \theta  , \hat M }$]{\includegraphics[width=0.30\hsize]{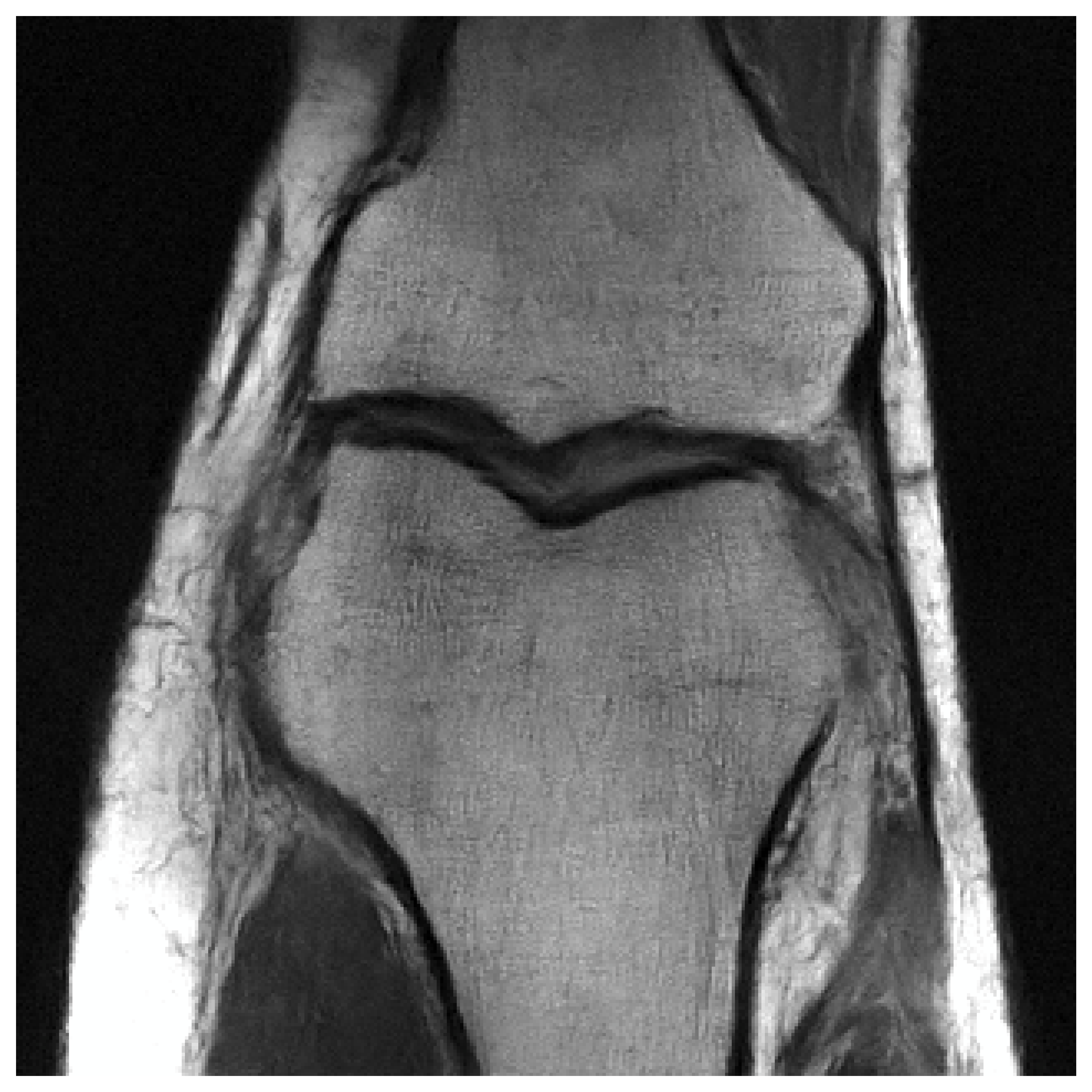}}
\subfloat[(c) Reference image]{\includegraphics[width=0.30\hsize]{figs/_gt.png}}
\caption{Posterior samples from our method with optimized design}\label{fig-post-samples}
\end{figure}

Both methods train an amortized posterior sampler that is sampled by passing Gaussian noise through the inverse network as in Equation \ref{eq-post-sampling}. We compare the quality of the posterior samples with the unoptimized mask in Figure
\ref{fig-post-samples-baseline} with the samples with the optimized mask in Figure \ref{fig-post-samples}. The posterior samples that used our optimized mask show sharper and more realistic features throughout the reconstruction.

By calculating the intrasample statistics of the posterior samples, we study the behaviour of the posterior mean and the posterior standard deviation. We calculate these statistics by taking posterior samples as calculated in Equation \ref{eq-post-sampling} then calculating the empirical statistics of the mean and standard deviation:
$$
\mathbb{E} \, p := \mathbb{E}_{\mathbf{x} \sim p(\mathbf{x}|\mathbf{y})} \left[ \mathbf{x}\right]
$$
$$ 
\mathbb{STDEV} \, p := \sqrt{  \mathbb{E}_{\mathbf{x} \sim p(\mathbf{x}|\mathbf{y})} \left[(\mathbf{x} - \mathbb{E} \, p  \right )^2]} .
$$ \label{eq-stdev}

\begin{figure}
\centering
\captionsetup[subfigure]{labelformat=empty}
\subfloat[(a) Baseline posterior mean: SSIM$=0.57$ ]{\includegraphics[width=0.30\hsize]{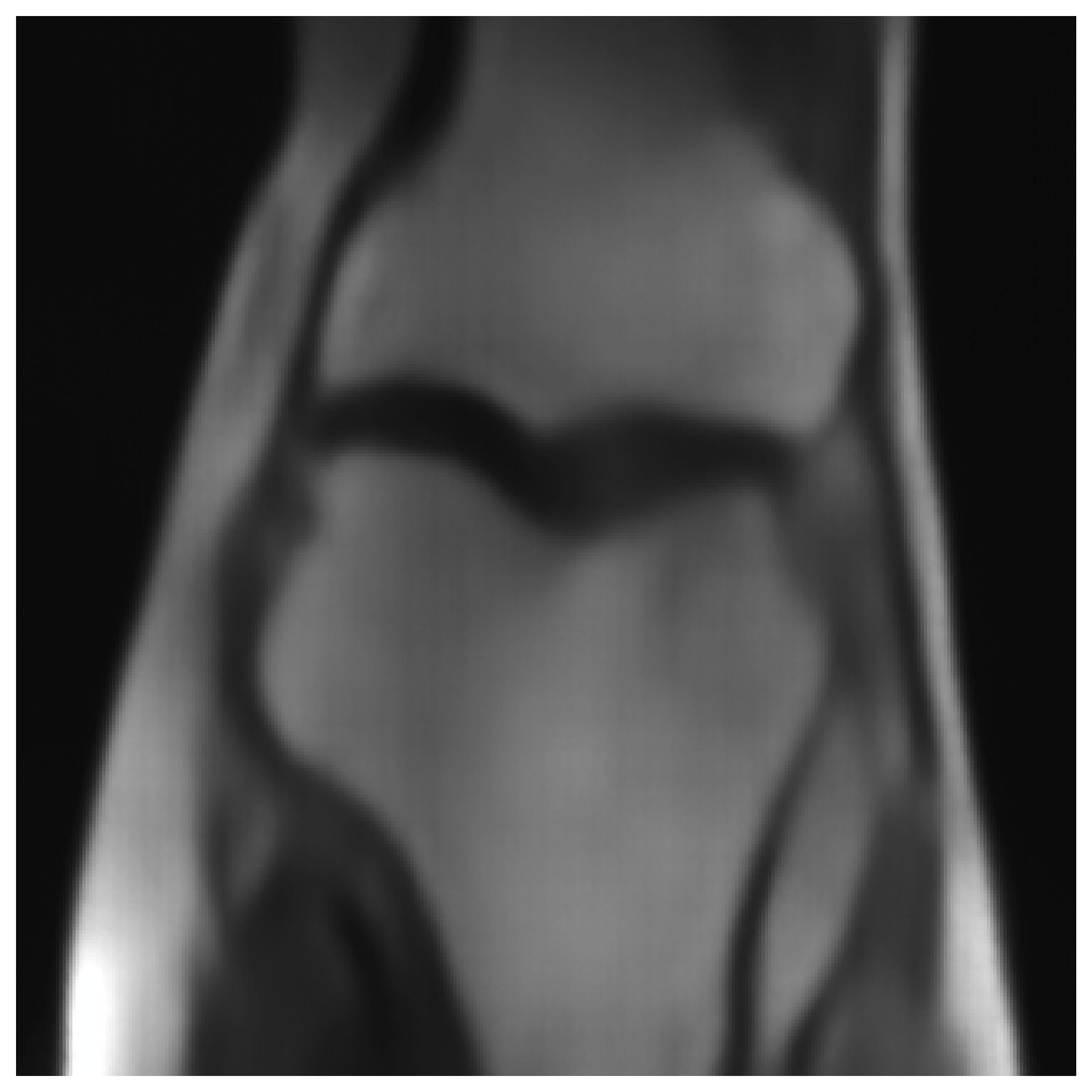}}
\subfloat[(b) Baseline posterior standard deviation]{\includegraphics[width=0.30\hsize]{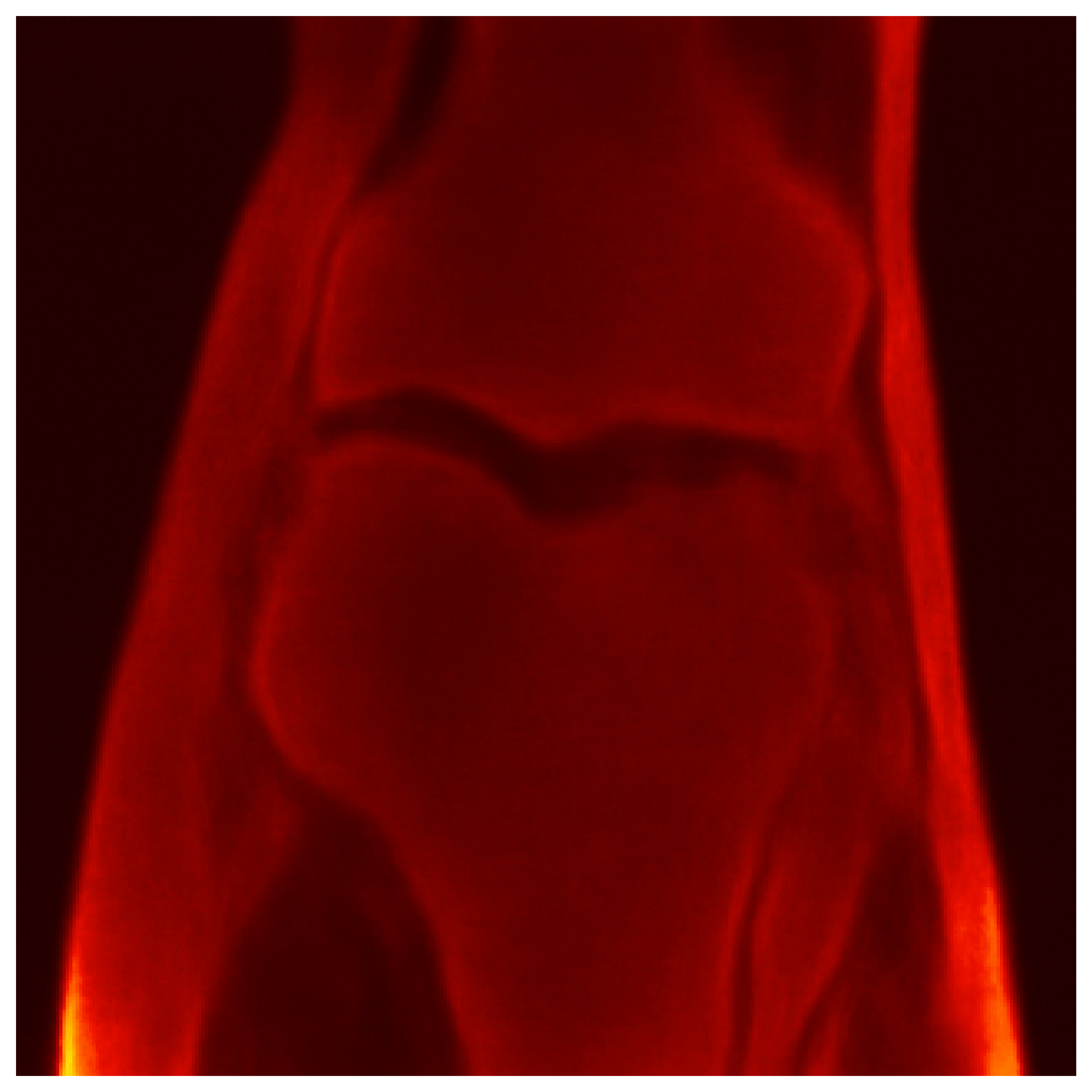}}
\subfloat[(c) Baseline error: NMSE$=0.1053$ ]{\includegraphics[width=0.30\hsize]{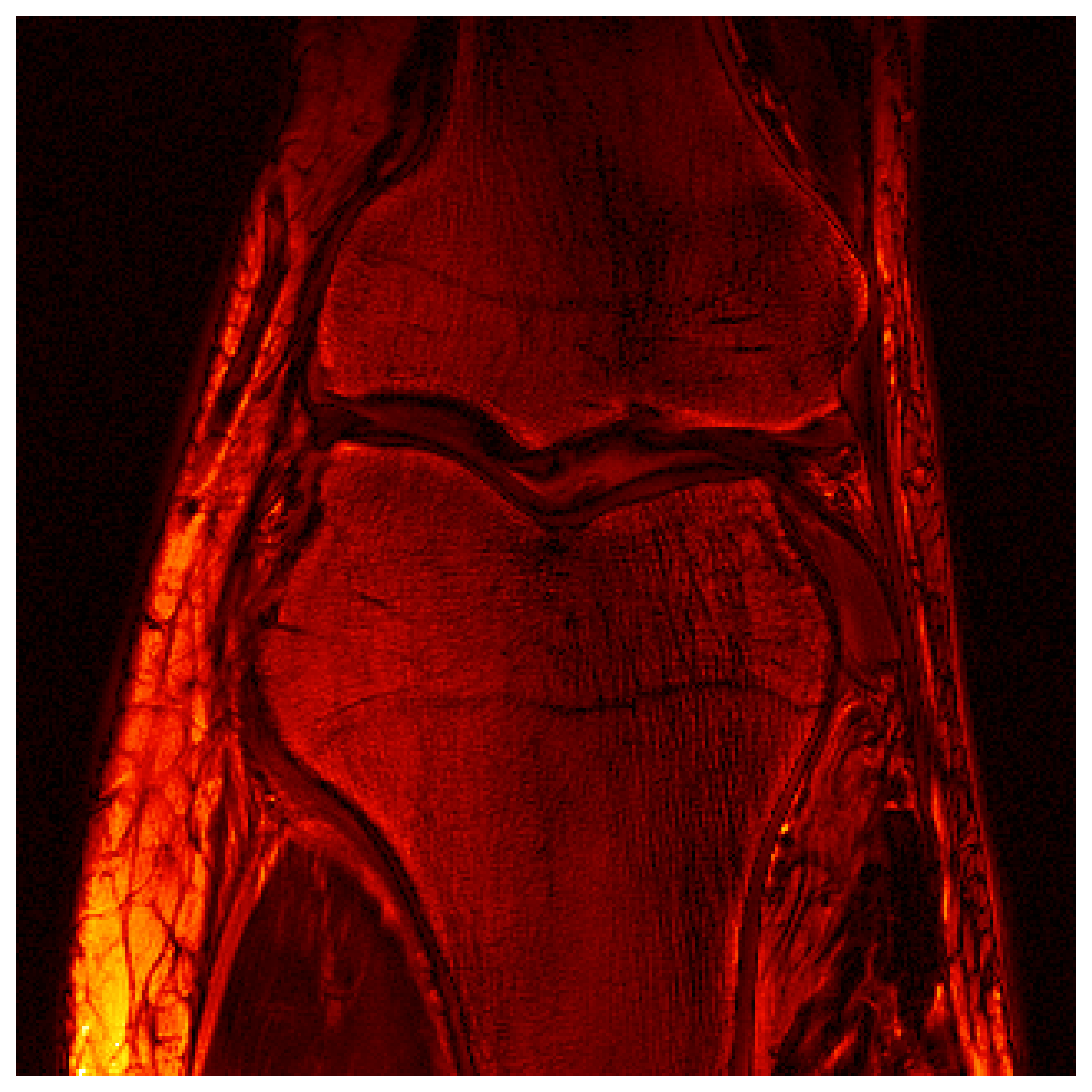}}
\\
\subfloat[(d) Our posterior mean: SSIM$=0.68$]{\includegraphics[width=0.30\hsize]{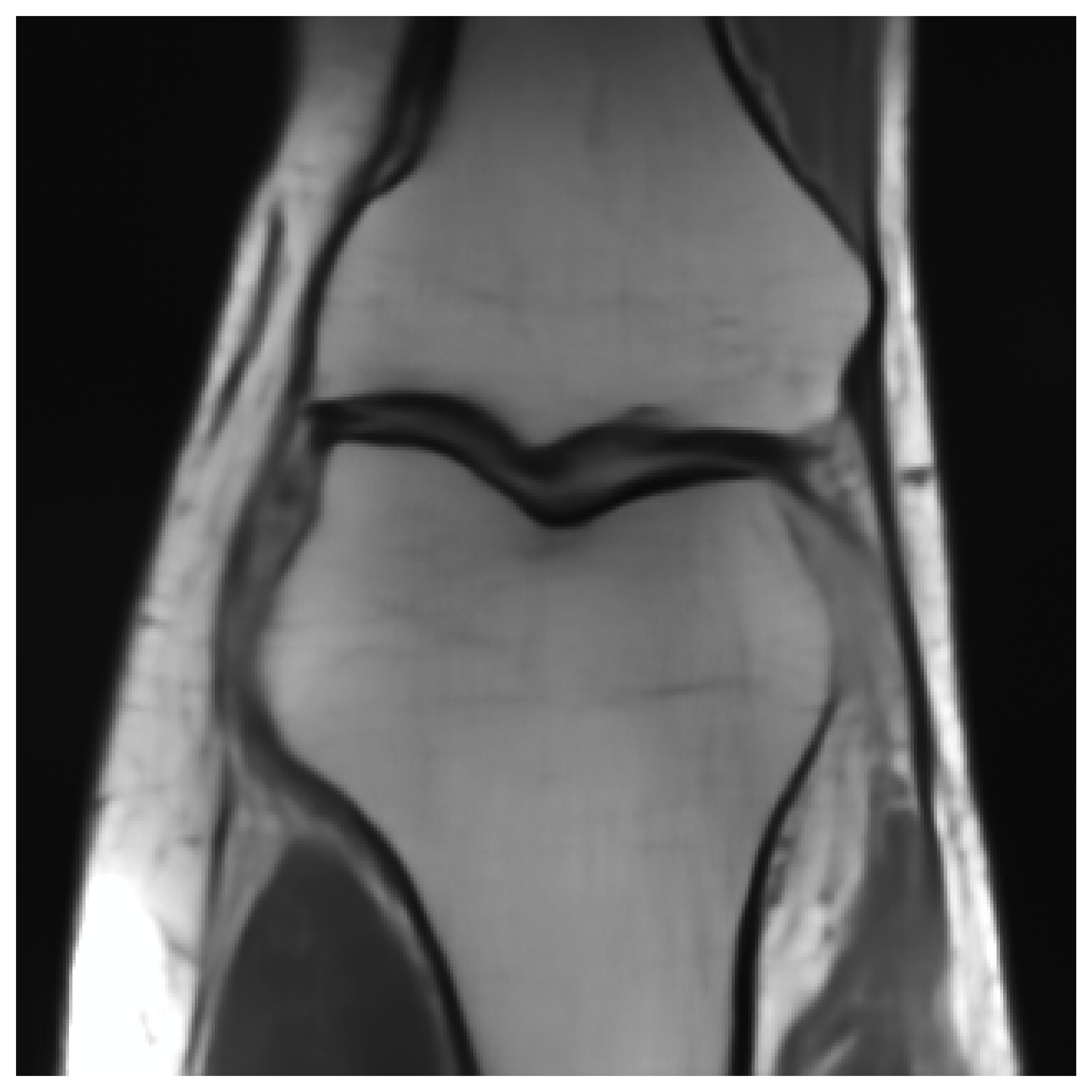}}
\subfloat[(e) Our posterior standard deviation]{\includegraphics[width=0.30\hsize]{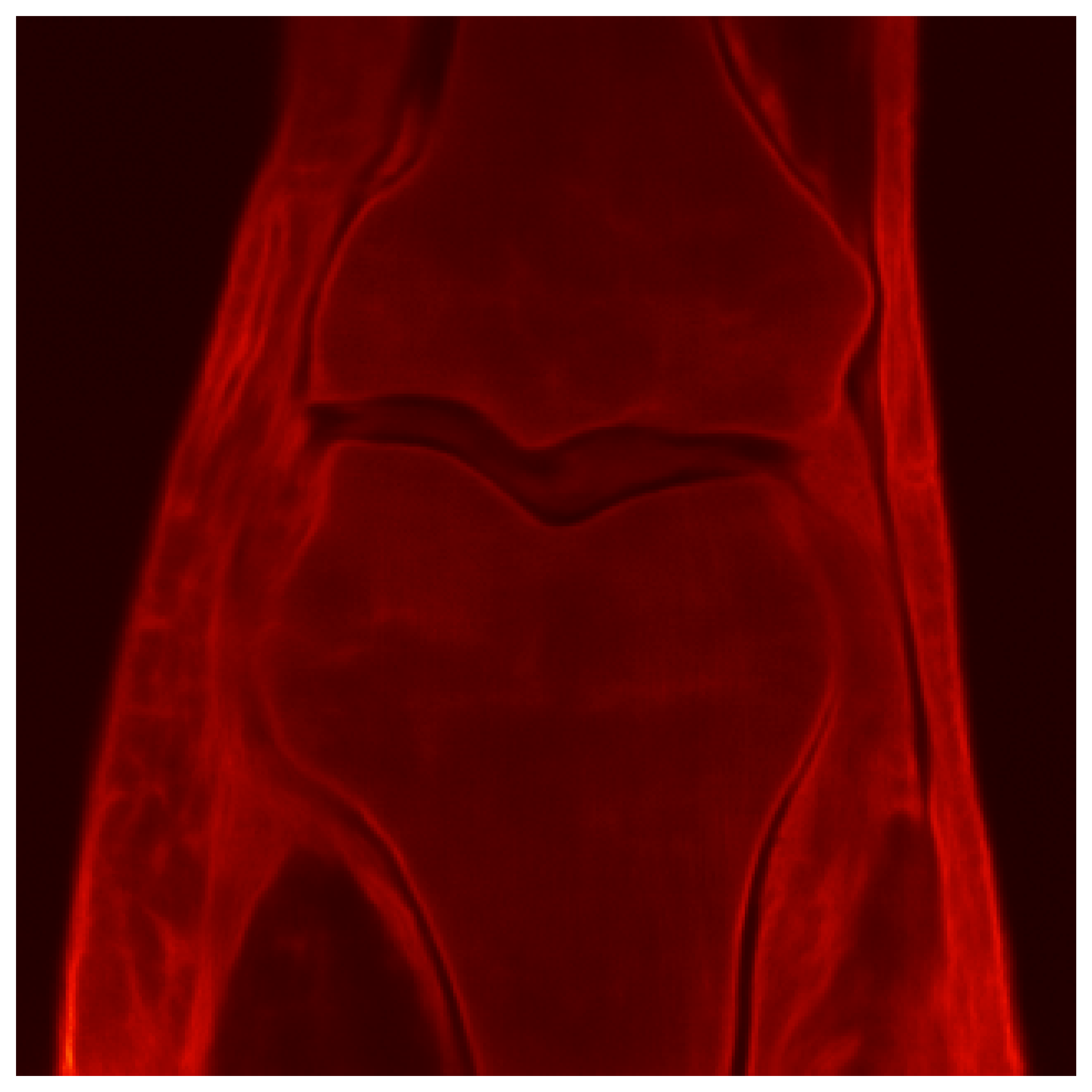}}
\subfloat[(f) Our error: NMSE$=0.0223$]{\includegraphics[width=0.30\hsize]{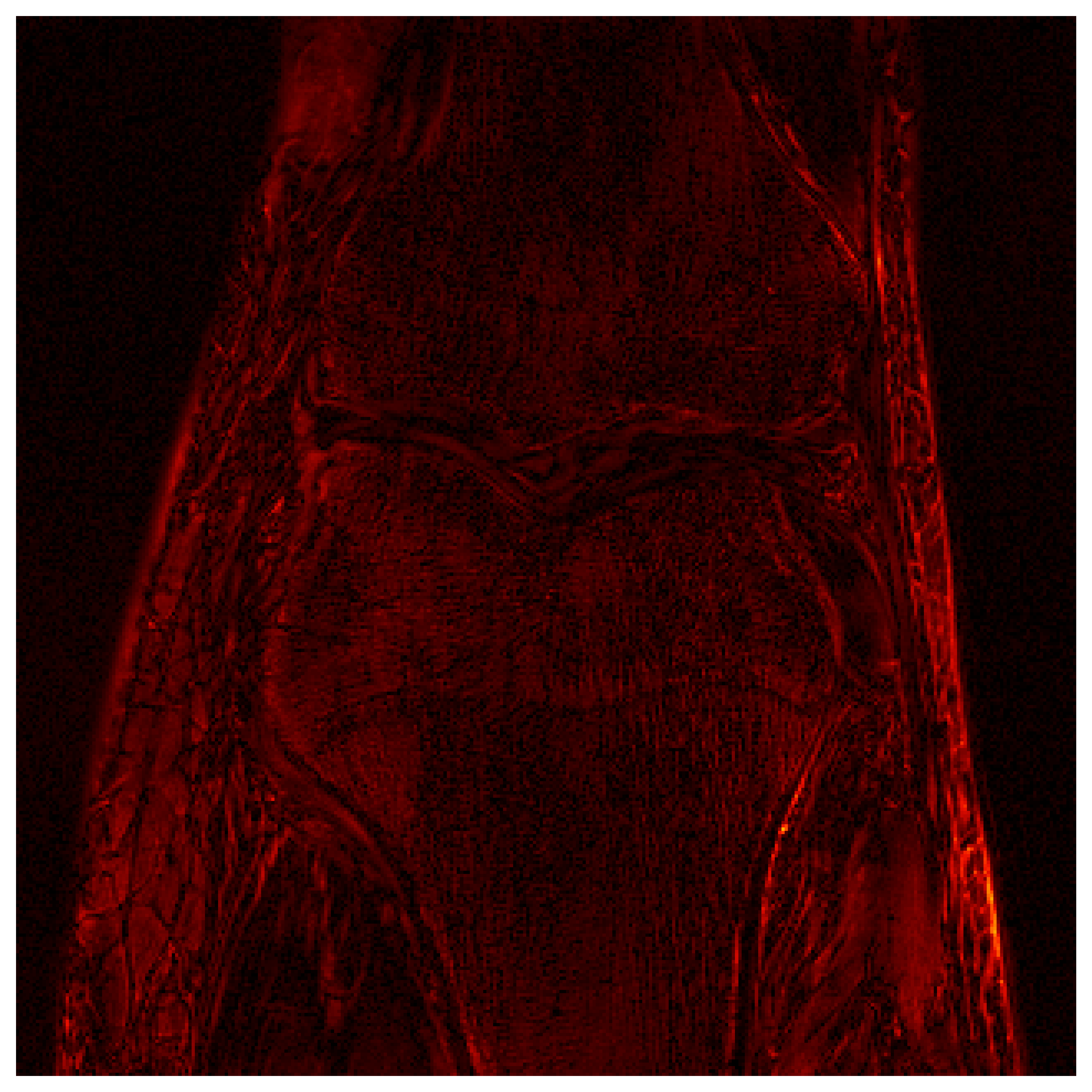}}

\caption{Pointwise statistics from the baseline compared to our method. }\label{fig-post-statistics}
\end{figure}

We plot these statistics for the same test sample in Figure \ref{fig-post-statistics}. We also plot the error of the posterior mean with respect to the reference image and note that for the test sample selected, our method shows less total uncertainty and also less error. 

\begin{figure}
\centering
\subfloat[Our optimized design reduces uncertainty]{\includegraphics[width=0.39\hsize]{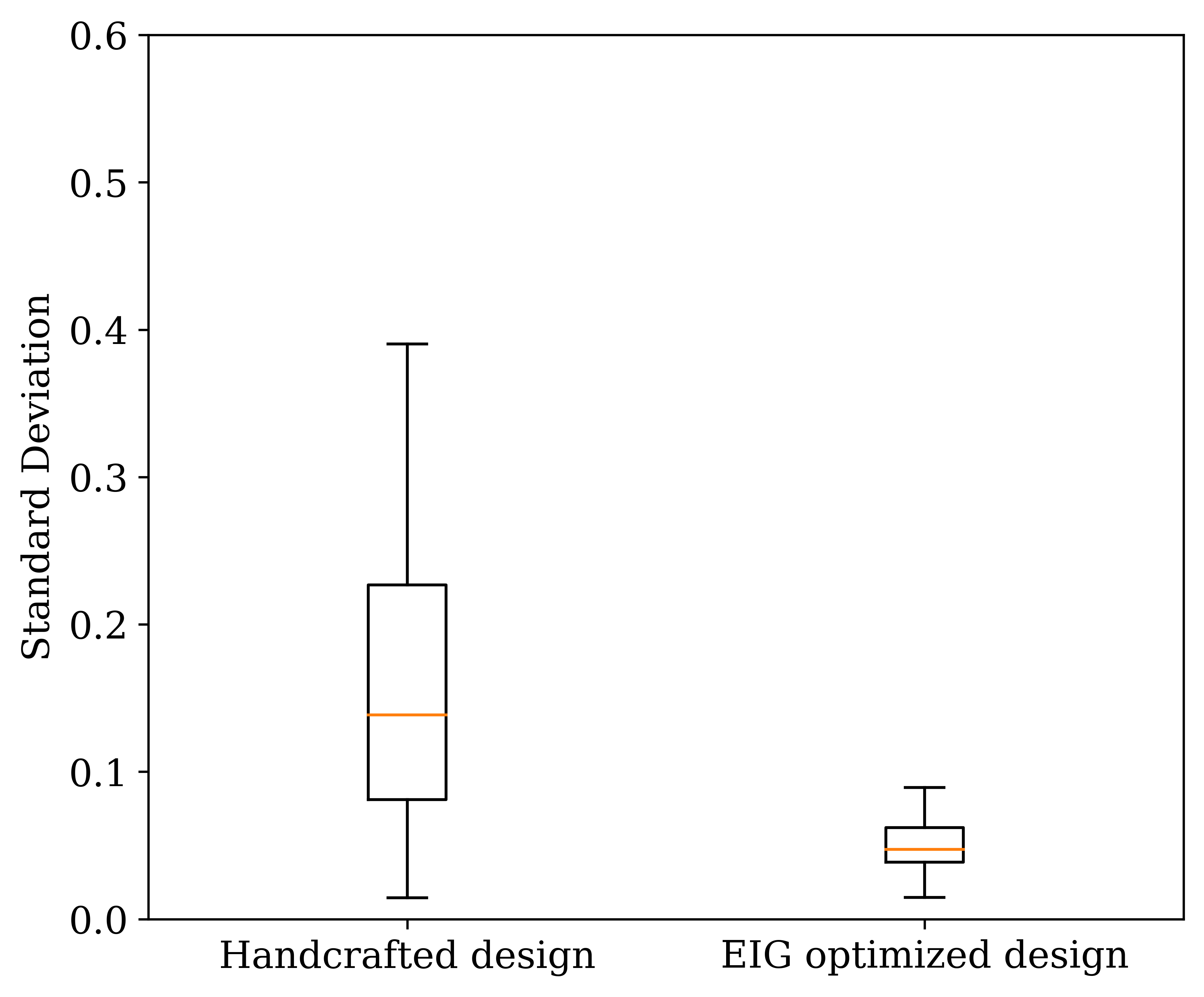}\label{fig-std-test}}
\subfloat[Reduced uncertainty leads to lower error]{\includegraphics[width=0.39\hsize]{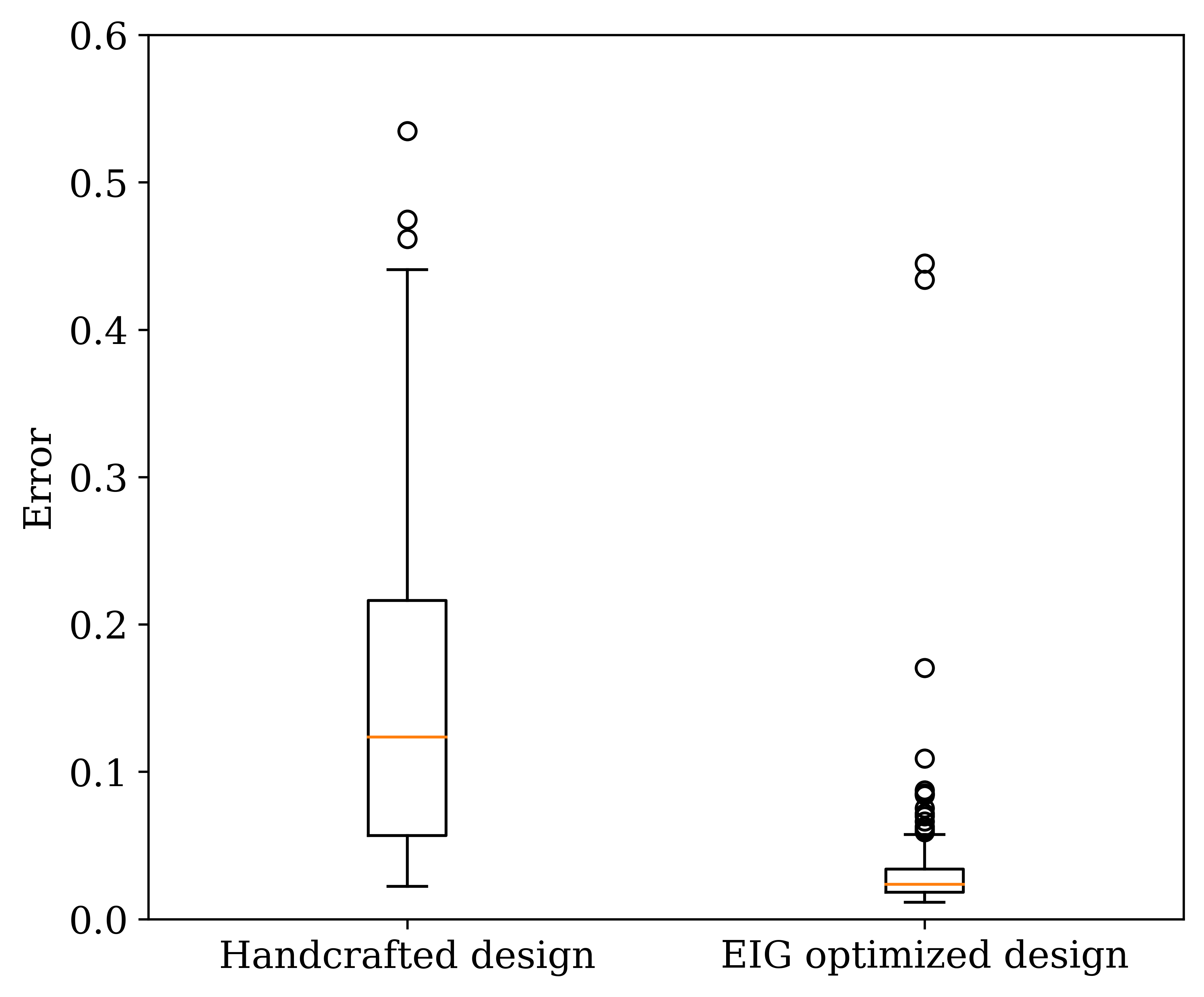}\label{fig-error-test}}
\caption{Comparing reduction in uncertainty achieved by our method.}\label{fig-tests }
\end{figure}

To quantify the gain in quality achieved by our optimal design as compared to the hand-crafted design, we calculate an image quality metric of normalized mean squared error (NMSE) between the posterior samples' mean and the ground truth over an unseen test set of $100$ samples. We also quantify the overall uncertainty of a certain posterior inference by calculating the normalized mean squared standard deviation. Since the $EIG$ is based on maximizing the posterior likelihood or equivalently lowering the posterior entropy then we expect our optimal design to lead to inference with less uncertainty. We summarize these results in Figure \ref{fig-std-test} showing that the uncertainty is consistently reduced by our method over the test set and furthermore that this translates to overall less error in the reconstruction given by the posterior mean as shown in Figure \ref{fig-error-test}.

\section{Related work}\label{Related}
We briefly outline similar work to this manuscript. The authors \cite{wu2023fast} use a pretraining phase that allows to efficiently compute the EIG based on a computed MAP approximation then efficient calculations of the posterior covariance. Given this EIG estimate, they proposed a set of greedy algorithms can be used to find demonstrably optimal EIG designs as compared to baselines. Similar to our work with normalizing flows,  \cite{kennamer2023design}, used the theory laid out in \cite{foster2020unified} to train an amortized conditional normalizing flow that works as an approximation to the $EIG$ and then shows that the $EIG$ is accurate as compared to a known analytical $EIG$. Although our methods are similar, we also learn the design jointly during normalizing flows training. 

In the medical field, \cite{wang2023sequential} trained a policy guided agent paired with a high quality reconstruction process to find optimal selections of measurement angles in sequential CT imaging. Specifically for MRI \cite{helin2022edge} showed that particular qualities such as edge promotion can be incorporated into a Bayesian experimental design framework. 

\section{Conclusions}\label{conclusions}

We have demonstrated the implementation of Bayesian experimental design on a realistic medical imaging problem. Our method relies on the exact likelihood density evaluation of normalizing flows leading to a simple method to jointly learn variational inference parameters and experimental design parameters. On top of this, the invertible architecture of normalizing flows enabled training on a large-scale imaging problem in MRI. Due to an absence of previous literature on solving this experimental design problem from a Bayesian perspective, we made compared our method with an equivalent Bayesian approach that does not use experimental design. Our experiments show gains in two downstream metrics: the reduction of the uncertainty in the posterior inference and the quality of the posterior samples' mean as measured with image quality metrics. 

\section{Acknowledgement}\label{acknowledgement}

This research was carried out with the support of Georgia Research Alliance and partners of the ML4Seismic Center. 

\bibliography{paper}

\end{document}